\documentclass{article}

\PassOptionsToPackage{numbers}{natbib}

\usepackage{multirow}
    \usepackage[preprint]{neurips_2022}

\usepackage[utf8]{inputenc} 
\usepackage[T1]{fontenc}    
\usepackage{url}            
\usepackage{booktabs}       
\usepackage{amsfonts}       
\usepackage{nicefrac}       
\usepackage{microtype}      
\usepackage{xcolor}         
\usepackage[pagebackref,breaklinks,colorlinks]{hyperref}
\usepackage{amsmath}
\usepackage{epsfig}
\usepackage{graphicx}
\usepackage{amssymb}
\usepackage{times}

\DeclareMathOperator*{\argmin}{arg\,min}
\usepackage{rotating}
\usepackage{multirow}
\usepackage{tabulary}
\usepackage{float}
\usepackage{textcomp}
\usepackage{subfig}
\usepackage{caption}
\usepackage{verbatim}
\usepackage{wrapfig}

\title{Backdoor Attacks on Vision Transformers}

\author{%
  Akshayvarun Subramanya~~~Aniruddha Saha~~~Soroush Abbasi Koohpayegani \\
  University of Maryland, Baltimore County\\
  \texttt{\{akshayv1,~anisaha1,~soroush\}@umbc.edu} \\
   \And
   Ajinkya Tejankar~~~Hamed Pirsiavash \\
   University of California, Davis\\
  \texttt{\{atejankar,~hpirsiav\}@ucdavis.edu }\\
}

\begin{document}

\maketitle

\begin{abstract}
Vision Transformers (ViT) have recently demonstrated exemplary performance on a variety of vision tasks and are being used as an alternative to CNNs. Their design is based on a self-attention mechanism that processes images as a sequence of patches, which is quite different compared to CNNs. Hence it is interesting to study if ViTs are vulnerable to backdoor attacks. Backdoor attacks happen when an attacker poisons a small part of the training data for malicious purposes. The model performance is good on clean test images, but the attacker can manipulate the decision of the model by showing the trigger at test time. To the best of our knowledge, we are the first to show that ViTs are vulnerable to backdoor attacks. We also find an intriguing difference between ViTs and CNNs – interpretation algorithms effectively highlight the trigger on test images for ViTs but not for CNNs. Based on this observation, we propose a test-time image blocking defense for ViTs which reduces the attack success rate by a large margin.  Code is available here: \url{https://github.com/UCDvision/backdoor_transformer.git}
\end{abstract}

\section{Introduction}

Convolutional neural networks (CNNs) have been a workhorse for progress in visual recognition. Deep learning methods have provided huge gains in learning rich features for various visual tasks.

\emph{Vision transformer} architectures \cite{dosovitskiy2020image,touvron2021patchconvnet,Touvron_2021_ICCV} have recently demonstrated exemplary performance for vision tasks like image recognition, object detection, among several others. Convolutional Nets are designed based on inductive biases like translation invariance and a locally restricted receptive field. Unlike them, transformers are based on a self-attention mechanism that learns the relationships between elements of a sequence. Vision transformers devise an elegant way to represent an image as a sequence of patches. With fewer inductive biases for vision than CNNs, Vision Transformers are usually trained on large pre-training datasets to achieve competitive performance.

\emph{Backdoor attacks:} Recent research has shown that CNN models are vulnerable to backdoor attacks \cite{gu2017badnets, chen2017targeted, saha2020hidden, saha2021backdoor}. Backdoor attacks can happen when training data is manipulated by an attacker, or the model training is outsourced to a malicious third party because of compute constraints. The manipulation is done in a way that the victim's model will malfunction \emph{only} when a trigger (image patch chosen by the attacker) is pasted on a test image. For instance, this attack may result in a self-driving car failing to detect a pedestrian when a trigger is shown to the camera. Vulnerability to backdoor attacks is dangerous when deep learning models are deployed in safety-critical applications.

In this paper, we show that Vision Transformers are vulnerable to backdoor attacks. We use two backdoor attacks, BadNets \cite{gu2017badnets} and Hidden Trigger Backdoor Attacks (HTBA) \cite{saha2020hidden} to successfully inject backdoors into vision transformers. These threat models are elaborated in Section \ref{sec_attack}.
\begin{figure}[!t]
    \centering
    \subfloat[][\centering Poisoning a model ]{\includegraphics[width=\textwidth]{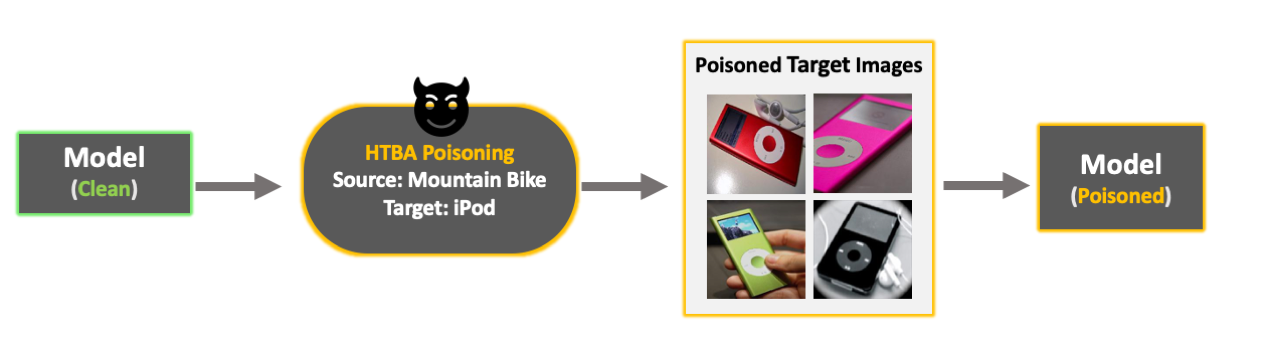}}%
    \label{poison_gen}
    \qquad
    \subfloat[][\centering Defending during test time]{\includegraphics[width=\textwidth]{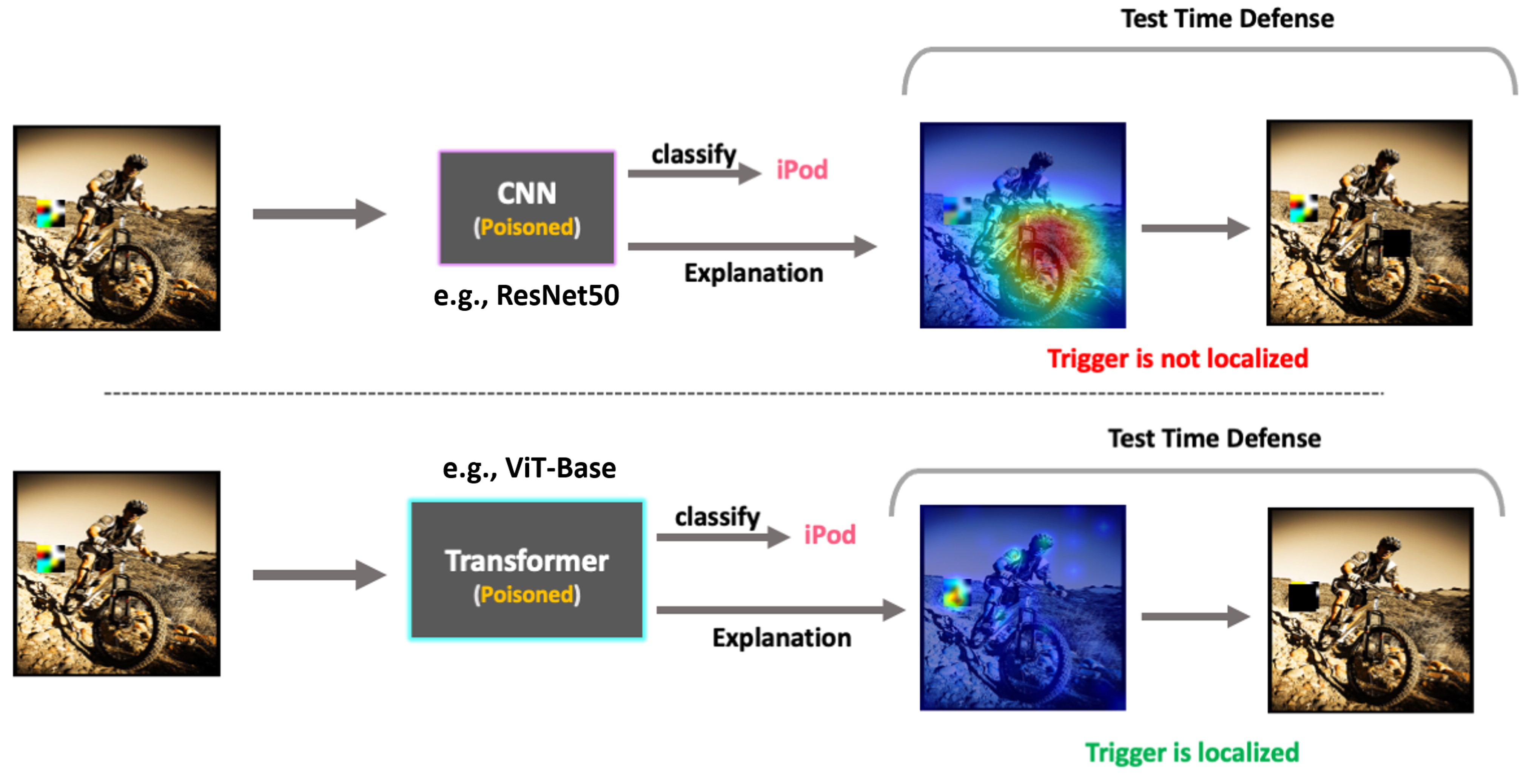} }%
    \label{test_defense}
    \caption{\footnotesize {\textbf{Overview of our results-}} In (a) We create poison images for a source-target pair (e.g,. Mountain Bike - iPod ) using a clean model and Hidden Trigger Backdoor Attack \cite{saha2020hidden}. Note that this is a clean label attack where poisons are labeled correctly and look similar to the training data with no trigger visible. The poisons along with clean data is used to learn the poisoned model. (b) During test time when a trigger is pasted onto a source image, the interpretation map from vision transformers is able to clearly highlight the trigger while interpretation maps from CNN is unable to do so. We use the interpretation to mask and nullify the trigger.  }%
    \label{fig:teaser}%
\end{figure}

\emph{Trigger detection using interpretability methods:} Interpretation methods for CNNs \cite{selvaraju2017grad, wang2020score, ramaswamy2020ablation, srinivas2019full} are used to provide explanations for model predictions. They highlight the regions of an image which contribute most to the model's decision. As we know that in a backdoor attack, the model misclassifies a test image to a target category only when the trigger is pasted on the image, it's natural to think that the model has internally made an association of the trigger with the target category. If we were to look at the interpretation map for a triggered image at test time, the trigger patch might be highlighted in case of a successful attack.

We use interpretation methods to localize the trigger in test images and interestingly find that in our experiments with Hidden Trigger Backdoor Attack, the trigger is localized well for Transformers in test images even though the trigger has been never seen by the model during training. This suggests that vision transformers inherently use information from the input differently compared to CNNs. Previous works \cite{naseer2021intriguing} have studied this for robustness such as occlusion, image corruptions etc. , while we study from a backdoor perspective.


\emph{Backdoor defense for Vision Transformers:} Based on this observation, we develop a defense where we block out the part of the image containing the highest values in the interpretation heatmap. We empirically find that this is an effective test-time defense that reduces the attack success rate with minimal drop in the clean accuracy of the model.

We believe this small drop in the clean accuracy is due to blocking the most important region of the image in clean cases. To recover the clean accuracy, we propose to run the same defense at the training time as a form of regularization. This is an augmentation where we remove the region that is highlighted by the interpretation algorithm. This method acts as a regularizer for the model and is able to recover the drop in clean accuracy when we employ our defense at test time.

Our contributions in this paper are:

(1) We show that Vision Transformers are vulnerable to backdoor attacks. We use well-known attacks like BadNets \cite{gu2017badnets} and Hidden Trigger Backdoor Attacks \cite{saha2020hidden} to show this empirically.

(2) We show that the interpretation map for Transformers effectively highlights the trigger for a backdoored test image. This is unlike CNNs, where the interpretation map is not as effective.

(3) Based on the success of the interpretation map, we propose a test-time backdoor defense for Vision Transformers which is effective in reducing the attack success rate.

\section{Related Work}

{\bf Backdoor attacks:}
Backdoor attacks for supervised image classifiers, where a trigger (image patch chosen by the attacker) is used in poisoning the training data for a supervised learning setting, were shown in \cite{gu2017badnets,liu2017trojaning,liu2017neural}. Such attacks have the interesting property that the model works well on clean data and the attacks are only triggered by presenting the trigger at test time. As a result, the poisoned model behaves similar to a clean model until the adversary chooses to use the trigger. Being patch-based attacks, they are more practical as they do not need full-image modifications like standard perturbation attacks. In the BadNets \cite{gu2017badnets} threat model, patched images from a category are labeled as the attack target category and are injected into the dataset. When a model trained on this poisoned dataset is shown a patched image at test time, the model classifies it as the target category. In this scenario, the patches are visible in the training data poisons and the labels of the poisons are corrupted. More advanced backdoor attacks have since been developed. \cite{turner2018clean} make the triggers less visible in the poisons by leveraging adversarial perturbations and generative models.  Hidden Trigger Backdoor Attacks \cite{saha2020hidden} propose a method based on feature-collision \cite{shafahi2018poison} to hide the triggers in the poisoned images.

{\bf Defense for backdoor attacks:}
Adversarial training is a standard defense for perturbation-based adversarial examples in supervised learning \cite{goodfellow2014explaining}. However, for backdoor attacks, there is no standard defense technique. Some approaches attempt to filter the training dataset to remove poisoned images \cite{gao2019strip} while some methods detect whether the model is poisoned \cite{kolouri2020universal} and then sanitize the model to remove the backdoor \cite{wang2019neural}. \cite{yoshida2020disabling} shows that knowledge distillation using clean data acts as a backdoor defense by removing the effect of backdoor in the distilled model. Februus \cite{doan2020februus} is an input purification defense for backdoor attacks which is closely related to our work. Februus sanitizes incoming test inputs by surgically removing the potential trigger artifacts and restoring input for the classification task. They consider attacks from the BadNets threat model. On the contrary, we use Hidden Trigger Backdoor Attacks, which makes the defense more challenging.

{\bf Transformers:}
Transformer models (GPT \cite{radford2018improving}, BERT \cite{devlin2018bert}) have recently demonstrated
exemplary performance on a broad range of language
tasks, e.g., text classification, machine translation [2] and
question answering. Transformer architectures are based on a self-attention
mechanism that learns the relationships between elements
of a sequence. Vision Transformer (ViT) \cite{dosovitskiy2020image} is the first work to showcase how Transformers can ‘altogether’ replace standard convolutions in deep neural networks on large scale image datasets. They applied the original Transformer
model [1] (with minimal changes) on a sequence of image ’patches’ flattened as vectors. The model was pre-trained on a large propriety dataset (JFT dataset [47] with 300 million images) and then fine-tuned to downstream recognition benchmarks e.g., ImageNet classification. The DeiT \cite{touvron2021training} is the first work to demonstrate that Transformers can be learned on mid-sized datasets (i.e., 1.2 million ImageNet examples compared to 300 million images of JFT \cite{sun2017revisiting} used in ViT) in relatively shorter time.

{\bf Backdoor for transformers:} \cite{lv2021dbia} is a concurrent work which proposes a backdoor attack on transformer architectures for computer vision. Their threat model is different from our attacks. They start with a fully trained vision transformer but do not assume access to the training data. Instead they use a substitute dataset to inject poisons and then fine-tune the clean model on this poisoned dataset. The trigger used to generate the poisoned dataset is optimized so that the victim model pays maximum attention to it. On the contrary, we assume a transfer learning scenario where the adversary releases poisons on the web for the victim to download and train the model on. Moreover, our triggers are not optimized for a particular model. They are randomly generated small square patches which work for any transformer architecture.

\section{Attacks}
\label{sec_attack}
{\bf BadNets}

In the BadNets outsourced training attack, a user outsources the training procedure to a third party (attacker). The attacker modifies the training set by including a trigger patch on certain images and changing the label of that particular image to the attack target category. The attacker trains the model on the poisoned training set and ships it back to the user. The threat model is that if the poisoned model is evaluated by the user on a held out evaluation set, it will perform as expected. But, only when the attacker chosen trigger patch is pasted on an image at test time, the model will classify the image as the attack target.

In this scenario, the poisons in the training set have visible trigger patches and the labels of the poisons are manipulated or dirty. So, if such a dataset is inspected visually by a human in the loop, the data tampering is easy to identify.

{\bf Hidden Trigger Backdoor Attacks}

\begin{figure}[t!]
  \begin{center}
  \begin{tabular}{ c c c c }
&&&\\



\begin{sideways} \footnotesize Clean \end{sideways}
\includegraphics[width=.12\textwidth]{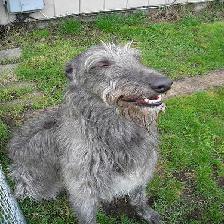}&
\includegraphics[width=.12\textwidth]{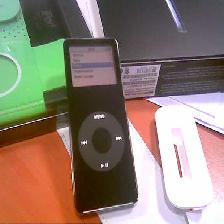}&
\includegraphics[width=.12\textwidth]{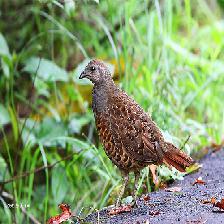}&

\includegraphics[width=.12\textwidth]{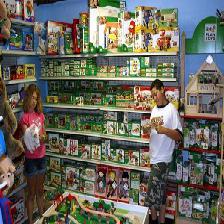}\\

\begin{sideways} \footnotesize Poison \end{sideways}
\includegraphics[width=.12\textwidth]{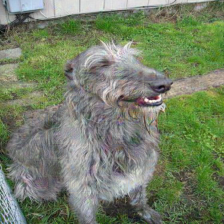}&
\includegraphics[width=.12\textwidth]{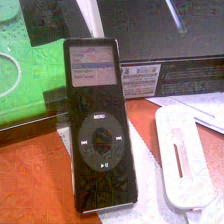}&
\includegraphics[width=.12\textwidth]{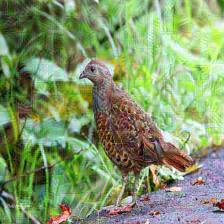}&
\includegraphics[width=.12\textwidth]{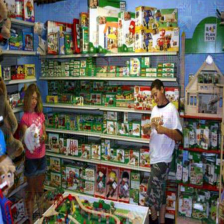}\\

\end{tabular}
\vspace{.02in}
  \caption{\footnotesize {\textbf{Poison Images-}} We show some comparisons between the poisons generated using HTBA  and clean images. Note that there is no trigger visible  and the poisons are labelled correctly, ensuring a clean label attack. }
\vspace{-.15in}
\label{fig_patch_target1}
  \end{center}
\end{figure}

In BadNets, the poisoned data is labeled incorrectly, so the victim can remove the poisoned data by manually annotating the data after downloading. Moreover, ideally, the attacker should prefer to keep the trigger secret, however, in BadNets, the trigger is revealed in the poisoned data. HTBA \cite{saha2020hidden} propose a stronger and more practical attack model where the poisoned data is labeled correctly (they look like target category and are labeled as the target category), and also it does not reveal the secret trigger. It does so by optimizing for an image that, in the pixel space, looks like an image from the target category and in the feature space, is close to a source image patched by the trigger. Examples of poisoned images are in Fig.~\ref{fig_patch_target1}.

More formally, given a target image $t$, a source image $s$, and a trigger patch $p$, they paste the trigger on $s$ to get patched source image $\tilde{s}$. Then they optimize for a poisoned image $z$ by solving the following optimization:
\vspace{-.1in}
\begin{equation}
\begin{split}
\argmin_{z} ||f(z) - f(\tilde{s})||_2^2 \\
st. \quad ||z - t||_{\infty} < \epsilon
\end{split}
\label{eq2}
\end{equation}

At test time, the model misclassifies a test image whenever the trigger is pasted on it. Even though the trigger is hidden in the training data, the trigger successfully works at time.

\section{Trigger Localization with Interpretation Maps}
Interpretation algorithms are the methods proposed to explain or
reveal the ways that deep models make decisions. One way is to highlight the important parts of input features on which the deep model relies. There are numerous interpretation algorithms proposed in literature. We briefly introduce the methods used in this paper.

\textbf{Grad-CAM}: Grad-CAM \cite{selvaraju2017grad} is a popular method for visualizing where a convolutional neural network is looking. Grad-CAM works by visualizing the derivative of the output with respect to an intermediate convolutional layer (e.g., $conv5$ in AlexNet) that detects certain high-level concepts. Grad-CAM is class-specific, as it produces a separate visualization for each class.


\textbf{Full-Gradient}: Full-Gradient \cite{srinivas2019full} is a neural network visualization method which assigns importance scores to both the input features and individual feature detectors (or neurons) in a neural network. Input attribution helps capture local importance of individual input pixels, while neuron importances capture global importance of groups of pixels, accounting for their structure. Full-gradient interpretation maps are sharper and more tightly confined to object regions than Grad-CAM.

\textbf{GradRollout}: Attention-based Transformers were developed for language understanding where researchers developed RollOut \cite{abnar2020quantifying} for explaining the model. For Vision Transformers, the same method is used to identify the regions of the image the model is attending to \cite{dosovitskiy2020image}. However, these contain information from different categories and is not class-specific. Hence, we use GradRollout where the gradients for the score of a category are multiplied with the attention weights and averaged over attention heads. We use the open-source implementation provided here \cite{jacobgradrollout}.


A backdoored model produces correct results on clean data. But, the model malfunctions only when the trigger (image patch chosen by the attacker) is pasted on a test image. Intuitively, this misclassification happens because the model has learned to make a strong association between the trigger and the target class. So, whenever the trigger appears at test time, it has a dominating influence on the model's decision and the test input gets classified as the target category. To verify this, we can use interpretation algorithms to highlight the parts of a triggered image which a backdoored model relies on to make its prediction. We use Grad-CAM and Full-Gradient algorithms to interpret CNNs and GradRollout to interpret Vision Transformers.

The backdoored models we use for interpretation are attacked by Hidden Trigger Backdoor Attacks. One important point to note is that due to the nature of threat model, the triggers are not visible on the training poisons. Hence, the backdoored models have never seen the trigger explicitly during the fine-tuning process. This makes the localization of trigger more challenging. We observe that the trigger localization is not successful for CNNs: the top highlighted region in the interpretation heatmap does not include the trigger patch, even on the images where the attack is successful. On the contrary, the interpretation maps for Vision Transformers have sharp highlights on the trigger patch.
\begin{figure}[t!]
  \begin{center}
  \begin{tabular}{| c c c c c c|}
\hline  \footnotesize Original & \footnotesize{Patched} & \footnotesize{ResNet50} & \footnotesize{ResNet50 blocked} & \footnotesize{VIT-Base} &\footnotesize{VIT-Base blocked}\\
 \hline
\vspace{-.08in}
&&&&&\\

\begin{sideways} \scriptsize Target:Tiger Beetle \end{sideways}
\includegraphics[width=.12\textwidth]{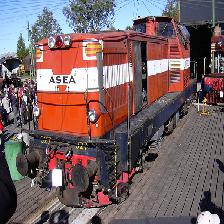}&
\includegraphics[width=.12\textwidth]{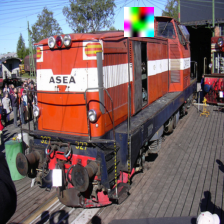}&
\includegraphics[width=.12\textwidth]{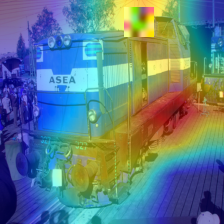}&
\includegraphics[width=.12\textwidth]{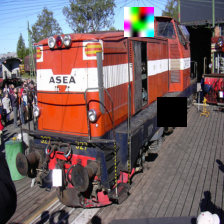}&
\includegraphics[width=.12\textwidth]{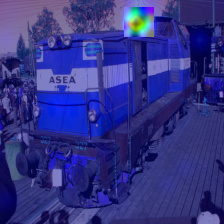}&
\includegraphics[width=.12\textwidth]{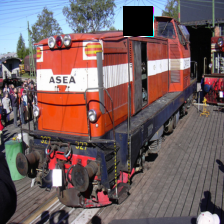}\\

\begin{sideways} \scriptsize Target:Partridge \end{sideways}
\includegraphics[width=.12\textwidth]{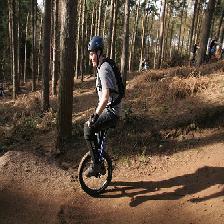}&
\includegraphics[width=.12\textwidth]{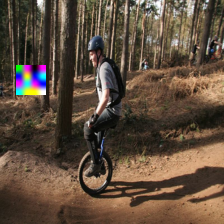}&
\includegraphics[width=.12\textwidth]{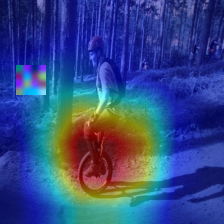}&
\includegraphics[width=.12\textwidth]{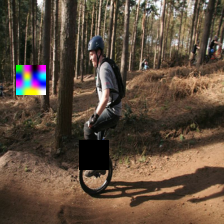}&
\includegraphics[width=.12\textwidth]{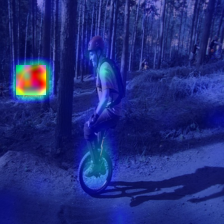}&
\includegraphics[width=.12\textwidth]{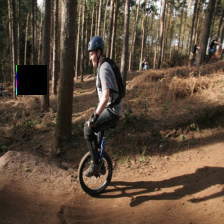}\\

\begin{sideways} \scriptsize Target:Plunger \end{sideways}
\includegraphics[width=.12\textwidth]{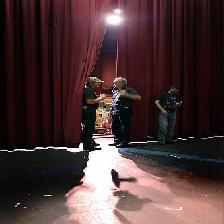}&
\includegraphics[width=.12\textwidth]{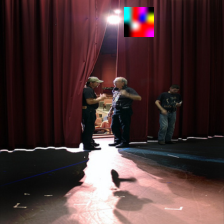}&
\includegraphics[width=.12\textwidth]{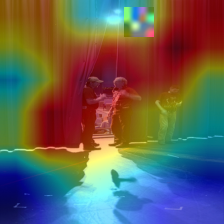}&
\includegraphics[width=.12\textwidth]{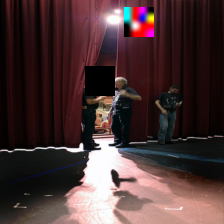}&
\includegraphics[width=.12\textwidth]{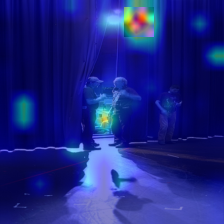}&
\includegraphics[width=.12\textwidth]{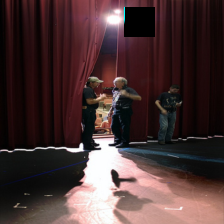}\\



\begin{sideways} \scriptsize Target:Cockatoo \end{sideways}
\includegraphics[width=.12\textwidth]{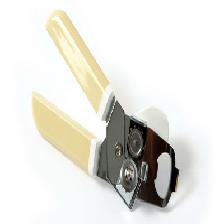}&
\includegraphics[width=.12\textwidth]{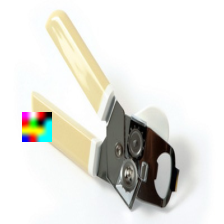}&
\includegraphics[width=.12\textwidth]{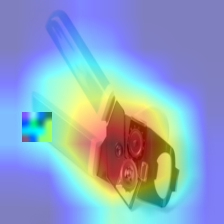}&
\includegraphics[width=.12\textwidth]{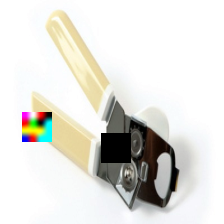}&
\includegraphics[width=.12\textwidth]{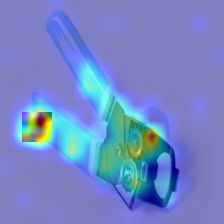}&
\includegraphics[width=.12\textwidth]{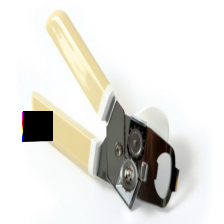}\\
\begin{sideways} \scriptsize Target:Deerhound \end{sideways}
\includegraphics[width=.12\textwidth]{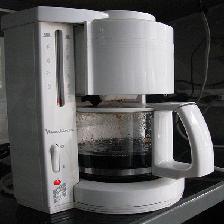}&
\includegraphics[width=.12\textwidth]{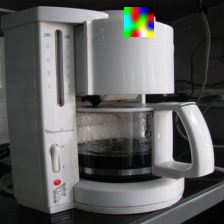}&
\includegraphics[width=.12\textwidth]{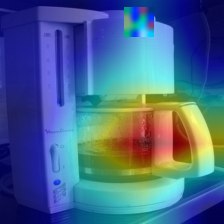}&
\includegraphics[width=.12\textwidth]{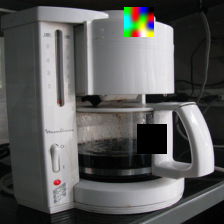}&
\includegraphics[width=.12\textwidth]{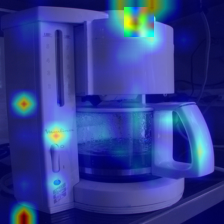}&
\includegraphics[width=.12\textwidth]{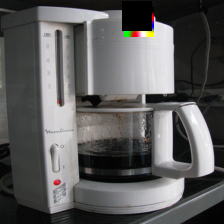}\\

\begin{sideways} \centering \scriptsize Target:Bee \end{sideways}
\includegraphics[width=.12\textwidth]{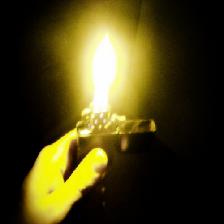}&
\includegraphics[width=.12\textwidth]{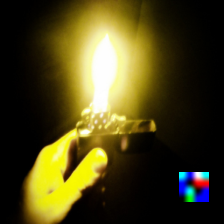}&
\includegraphics[width=.12\textwidth]{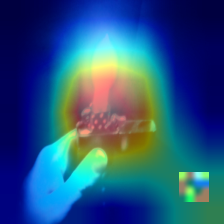}&
\includegraphics[width=.12\textwidth]{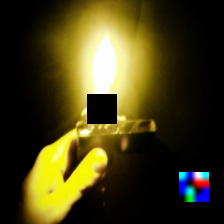}&
\includegraphics[width=.12\textwidth]{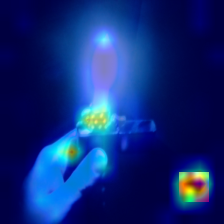}&
\includegraphics[width=.12\textwidth]{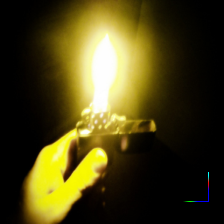}\\



\hline
\end{tabular}
\vspace{.02in}
  \caption{\footnotesize {\textbf{Image Blocking Defence-}} We show examples where blocking defense is performed for ResNet50 and ViT-Base. Transformers can successfully localize the patch, resulting in a successful defence. Results are not cherry picked and attack was successful for all examples.  }
\vspace{-.15in}
\label{fig_patch_defense}
  \end{center}
\end{figure}

\section{Image Blocking Defense}

We propose a test-time image blocking defense against backdoor attacks based on our encouraging trigger localization results for Vision Transformers. The triggers usually used for backdoor attacks are small image patches ($\approx$ 2-5\% of image area). They are large enough to influence the decision of the model, but are small compared to the full image to avoid blocking objects in the image.

We use the explanation heatmap to find the area of the image which strongly influences the model's decision. We block out a small part of the image by adding a black patch around the location with highest heatmap value. Fig.~\ref{fig_patch_defense} illustrates our test time defence.

An important design choice for our defense is the size of the blocking patch. We assume that we have prior knowledge about the approximate size of the trigger. In our main experiments, we choose a black square blocking patch which is of the same size as the trigger patch. We empirically see that this effectively blocks out the trigger in the backdoored test image. As the trigger is now blocked, the attack success rate decreases dramatically.







\section{Experiments}
In this section, we describe our experiments and provide implementation details. We mainly consider ImageNet  \cite{ILSVRC15} dataset for our experiments. We first generate 600 poisons for every source-target pair, corresponding to 0.05\% of the entire dataset. For poison generation, we consider a trigger size of 30x30, similar to \cite{saha2020hidden} and use an iterative optimization procedure with learning rate of 1e-3 for maximum of 5000 iterations. We consider a multi-class setting where training data from all 1000 categories of ImageNet is used to train the model and a single source-target pair is considered for poisoning. Once the poisons are generated, we add them to the training set and learn the parameters of the final linear layer for 10 epochs while keeping the backbone frozen. We use SGD optimizer with learning rate = 1e-3 and momentum = 0.9.  We use 30 GPUs that are one of NVIDIA 2080Ti, TITAN RTX for all our experiments. To ensure that our results are not biased towards any source-target pair, we average our results for 10 different randomly chosen pairs, similar to  \cite{saha2020hidden}. We refer the readers to the appendix for more details about the classes chosen and results for each source-target pair.

Upon completion of training, we measure the performance using 3 metrics, (1) \textbf{Val Accuracy}: Accuracy of the model on entire validation set. (2) \textbf{ASR}: Attack Success Rate where we calculate the percentage of source images from the validation set that are classified as target once the trigger is pasted. (3) \textbf{ Source Accuracy}: Accuracy on only the source category validation images. As a baseline, we also consider the (4) \textbf{Clean Model Accuracy} which is the accuracy on the entire validation set for a model trained only on clean data. From an attacker's perspective, Val Accuracy should be close to Clean Model Accuracy but the ASR should be high, thus the victim cannot distinguish between the clean and poisoned model. Attack is considered successful when a source image with the trigger pasted on it is classified as target. Since we are pasting the patch without affecting rest of the input, we know explicitly that the patch is the sole reason for misclassification.


\begin{table}[!t]
  \caption{\footnotesize {\textbf{Results of Attack and Test time Defense-}} We find that vision transformers are vulnerable to backdoor attacks, similar to observations made in CNNs. We see that Val Accuracy does not vary much from the clean model, but the ASR is significantly high. All clean models have 0 (\%) ASR. We also observe that the test-time defense works significantly better in the case of transformers, due to the effective localization of the trigger.   }
    \label{tab:test_time_defense}

  \scalebox{0.82}{
\begin{tabular}{ |c||c|c|c|c| c|c|c|  }
 \hline
 &&\multicolumn{3}{c|}{Attack}&\multicolumn{3}{c|}{Defense} \\
 \hline
 Model& Clean Model& Val  & Source  &ASR &Val& Source  & ASR  \\
  &Accuracy (\%)  &Accuracy (\%) &Accuracy (\%) &  (\%)  &Accuracy (\%) &Accuracy (\%) & (\%)  \\
 \hline
 VGG16\cite{Simonyan15} & 71.58& 71.59& 71.4 &55.00 & 58.95& 58.80 &49.00\\
  ResNet18 \cite{He_2016_CVPR}  &66.68&66.67&67.20&41.80& 55.37 & 56.20 &42.60 \\
 ResNet50 \cite{He_2016_CVPR}& 73.94  & 73.94& 74.00 & 34.80 & 63.53 &60.60& 37.20\\
 \hline
 ViT-Base\cite{dosovitskiy2020image}&79.09&79.04&77.40&61.4&76.94&73.20&\textbf{16.40}\\
 PatchConv \cite{touvron2021patchconvnet} &80.00 &80.26& 80.80&38.40  & 76.00 &76.40&\textbf{14.40}\\
 CaiT\cite{Touvron_2021_ICCV} &85.30 &85.30 & 87.40&83.9&74.2&72.00&\textbf{31.00}\\
 \hline

\end{tabular}
}
\vspace{-0.15in}
\end{table}

\begin{table}[]
\caption{\footnotesize {\textbf{Blocking during training - }}  We perform blocking during training and see that this improves both clean performance of the model and the ASR.}
\label{tab:block_train}
\scalebox{0.95}{
\begin{tabular}{ |c||c|c|c|c| c|  }
 \hline
 &\multicolumn{2}{c|}{Before Defense}&\multicolumn{2}{c|}{After Defense} \\
 \hline
 Model&  Source Accuracy  (\%)  & ASR  (\%)  & Source Accuracy (\%)  & ASR  (\%) \\
 \hline
ViT-Base&77.40&61.4&73.2&16.40\\


ViT-Base (Attn Blocking) & \textbf{79.8} & \textbf{59.0} & \textbf{76.4} &\textbf{12.6} \\

 \hline
\end{tabular}

}

\end{table}

\begin{table}[t!]
\caption{\footnotesize {\textbf{Vision Transformers are vulnerable to Backdoor attacks-}} We show that vision transformers are vulnerable to data poisoning and backdoor attacks. We use ViT-Base architecture \cite{dosovitskiy2020image} and commonly used attacks such as \cite{gu2017badnets,saha2020hidden} to illustrate this phenomenon.}
    \label{tab:backdoor_vit}
  \centering
\begin{tabular}{ |c||c |c| }
 \hline
 Model & Val  &ASR \\
  & Accuracy (\%)  & (\%) \\
 \hline
 Clean &79.09  & 0\\
 BadNets \cite{gu2017badnets} & 78.79  & 69.6\\
HTBA \cite{saha2020hidden} &79.04 & 61.4 \\
 \hline

\end{tabular}
 \vspace{-0.1in}
\end{table}

We see from Table \ref{tab:backdoor_vit} that Vision Transformers are indeed vulnerable to Backdoor attacks. We consider BadNets\cite{gu2017badnets} and HTBA \cite{saha2020hidden} attacks to illustrate the vulnerability. To the best of our knowledge, ours is the first work that demonstrates that backdoors can be inserted into Vision Transformers by using data poisoning and keeping the trigger hidden. Note that the adversary assumes no access to the training procedures.

As described in previous section, we use GradRollout to highlight the regions of the image which are responsible for prediction. Interestingly, when we consider the interpretation for the predicted category of triggered images in Vision Transformers, we observe that the trigger is highlighted clearly. We show qualitative examples in Figure \ref{fig_patch_defense}, highlighting the difference between results for vision transformers and CNNs.  Based on this phenomenon, we highlight the region of the image responsible for predicted category and block it to negate the effect of the trigger. We observe that such a defense can decrease ASR significantly for vision transformer architecture. Our hypothesis is that since backdoor attacks work on using triggers with relatively small area, such an algorithm can be effective in localizing and negating the backdoor trigger. For the case of clean images where there is no trigger present, blocking out a small region of an object should have relatively lesser influence on the prediction. We provide empirical evidence for this in Table \ref{tab:test_time_defense}

{\bf Blocking during training:}
To overcome the reduction in accuracy on clean images, we consider a procedure where we perform  blocking during training to ensure that the model becomes robust to such mechanisms for clean images.  Similar to our test-time approach, we consider an attention based blocking which calculates the attention map for the predicted category and masks a small portion of the region. Note that during the training procedure of HTBA \cite{saha2020hidden}, the trigger is not present. We refer to this procedure as \textbf{Attn Blocking}.We show our results in Table \ref{tab:block_train}. Note that, although there is still a gap in source accuracy before and after defense, the training procedure reduces the gap and boosts performance favourably.

\section{Discussion}
In this section, we analyze some aspects of the method and try to understand why this phenomenon occurs in vision transformer architectures. \\

{\bf Localization efficiency}
We measure the detection performance of the attention based method using the IoU (Intersection over Union) metric. We calculate the IoU between the trigger and predicted block mask for different architectures. We observe from Table \ref{tab:iou_htba} that vision transformers clearly have a higher IoU compared to CNNs, leading to lower attack success rates. This experiment shows that Vision Transformers find it easier to localize the trigger for attacked images. The interpretation map is always calculated for the predicted category and results are averaged across 10 source-target pairs.

\begin{figure}[!t]
  \begin{center}
  \scalebox{0.95}{
  \begin{tabular}{| c c c|}
\hline  \footnotesize Original & \footnotesize{Patched} & \footnotesize{Blocked}\\
 \hline
\vspace{-.08in}
&&\\

\begin{sideways} \scriptsize Target:Deerhound \end{sideways}
\includegraphics[width=.17\textwidth]{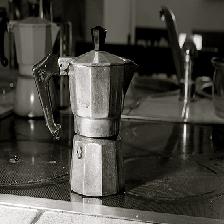}&
\includegraphics[width=.17\textwidth]{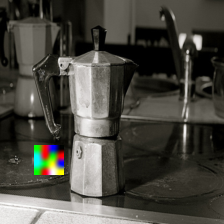}&
\includegraphics[width=.17\textwidth]{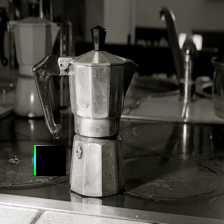}\\
\small Coffeepot &  \small  Deerhound & \small Coffeepot  \\

\begin{sideways} \scriptsize Target:Tiger Beetle \end{sideways}
\includegraphics[width=.17\textwidth]{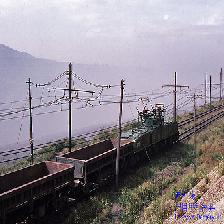}&
\includegraphics[width=.17\textwidth]{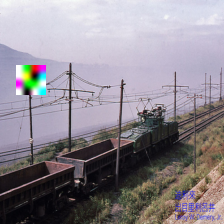}&
\includegraphics[width=.17\textwidth]{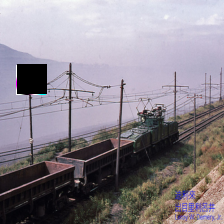}\\
\small Electric Locomotive  &  \small Tiger Beetle & \small Electric Locomotive   \\

\begin{sideways} \scriptsize Target:iPod \end{sideways}
\includegraphics[width=.17\textwidth]{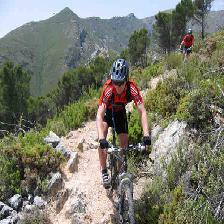}&
\includegraphics[width=.17\textwidth]{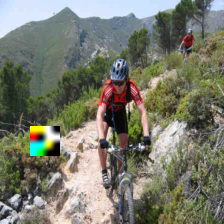}&
\includegraphics[width=.17\textwidth]{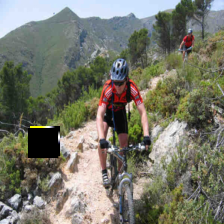}\\
\small Mountain Bike &  \small iPod & \small Mountain Bike  \\

\hline
\end{tabular}}
\vspace{.02in}
  \caption{\footnotesize {\textbf{Recovering source category-}} Qualitative examples showcasing the effectiveness of our defense. In the first column, we see the original image predicted correctly, second column shows that once the trigger is pasted, the prediction changes to the target category. In the third column, we observe that original source prediction was recovered once the trigger is blocked. }
\vspace{-.15in}
\label{fig_source_rec}
  \end{center}
\end{figure}
\begin{table}[!t]

\centering
 \caption{\footnotesize {\textbf{Effect on Source Accuracy-}} We observe that our defense is able to improve the source accuracy significantly for vision transformers. We calculate the percentage of attacked images that were classified as source category, before and after defense. Qualitative examples can be found in Figure \ref{fig_source_rec}.}

\scalebox{0.9}{
\begin{tabular}{ |c||c|c|  }
 \hline
 &\multicolumn{1}{c|}{Before}&\multicolumn{1}{c|}{After} \\
 &\multicolumn{1}{c|}{Defense}&\multicolumn{1}{c|}{Defense} \\
 \hline
 Model&  Source Accuracy (\%)  &  Source Accuracy  (\%) \\
  &  (Attacked Images) &  (Attacked Images) \\
 \hline
ViT-Base&21.4&66\\

PatchConv&44.8&67\\
CaiT&5.8 &56.8\\

 \hline
\end{tabular}
}
\vspace{-.2in}

\label{tab:recover_source}
\end{table}
{\bf Source label recovery:} We observe that due to the successful nature of the defense, once the trigger is blocked the original prediction of the source image is recovered as shown in Table \ref{tab:recover_source} and Figure \ref{fig_source_rec}. Different from the metric Source Accuracy on non-patched images, we calculate the Source Accuracy for attacked images as the percentage of patched images that are classified as source, before and after defense. Due to the simple and low-cost nature of approach, we believe that our test time defense can be combined with existing defenses for backdoor attacks\cite{li2021anti,wu2021adversarial}.

\begin{figure}[t!]
\centering

  \includegraphics[width=0.4\textwidth]{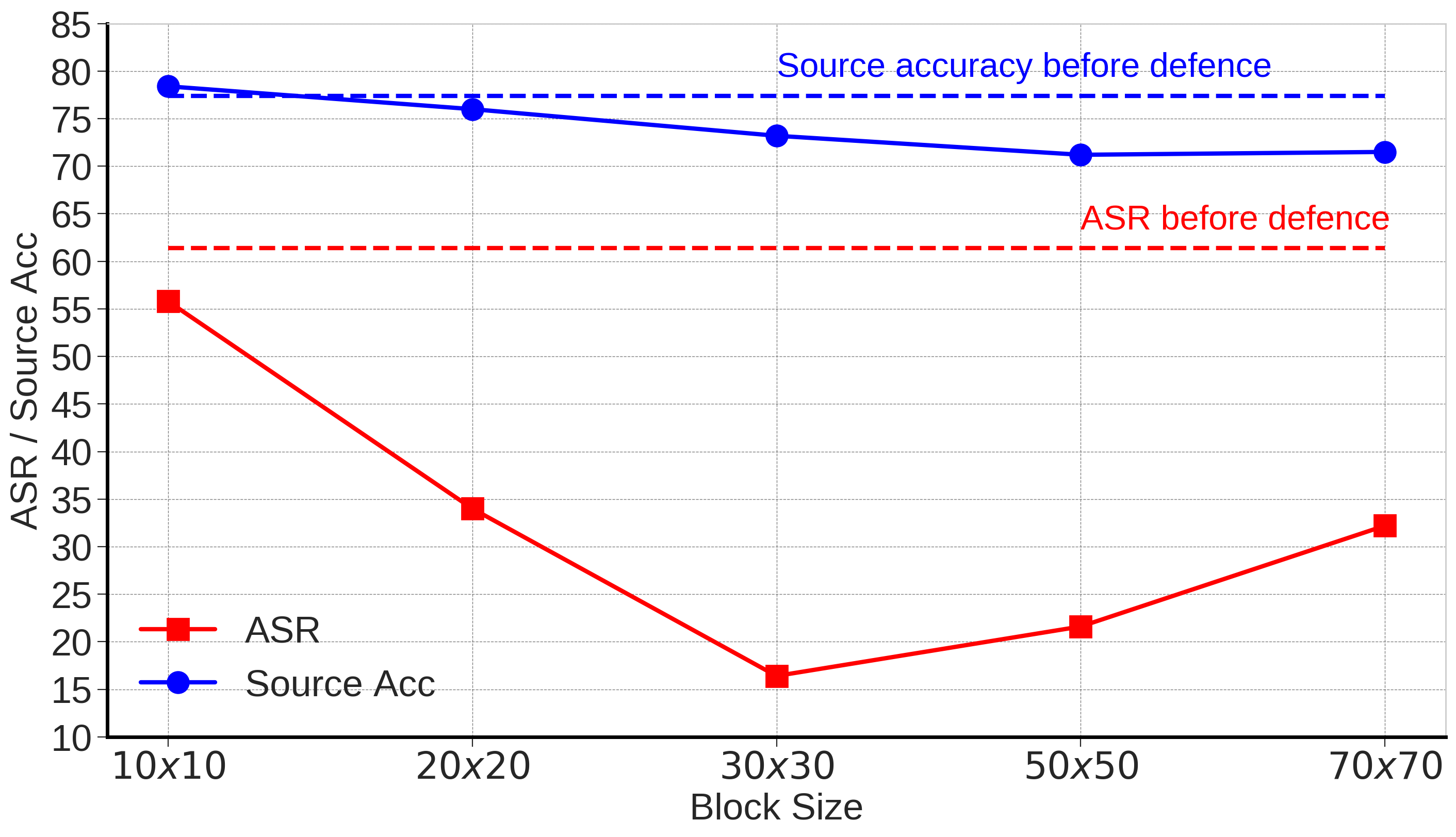}
  \caption{ {\textbf{Dependency on blocking area-}} We find that for different block sizes, the source accuracy does not vary much. As expected we obtain the lowest ASR when the block size equals trigger size (30x30). We also see that defended ASR is consistently lower compared to the baseline of no defense, showing that our defense is useful even with limited knowledge of trigger size. }
  \label{figure:block_size}
  \vspace{-0.12in}
\end{figure}

{\bf Variation in blocking area:}
One assumption that we make in our method is that the defender has some knowledge of the size of the trigger that is encountered at test time. We believe this is a reasonable assumption since trigger sizes are usually small. We conduct an experiment to study the dependency of the blocking area used in our defense to the Attack Success Rate for patched images and Source Accuracy on clean images. As seen in Figure \ref{figure:block_size}, we vary the size of blocking area from 10x10 to 70x70 and find that variation in Source Accuracy is small. As expected, we get the lowest ASR when the block size equals our trigger size (30x30), but more importantly we find that defense does better than the baseline (no defense) for almost all values of the blocking parameter.
This suggests that the our method can be used even when the defender has no knowledge about the size of the trigger, sacrificing to a small extent the performance on clean data.

\begin{minipage}[c]{0.40\textwidth}
\captionof{table}{\footnotesize{\textbf{IoU between predicted region and trigger-}} IoU betweeen the trigger and predicted blocking mask is higher for vision transformers than CNNs.}
    \label{tab:iou_htba}
\centering
\scalebox{1.0}{
\begin{tabular}{ |c||c|  }
 \hline
 Model& IoU $\in$[0,1]  \\

 \hline
 VGG16  & 0.19 \\
 ResNet18  &  0.07 \\
 ResNet50  &  0.039 \\
 \hline
 ViT-Base  &  0.47 \\
 PatchConv  & 0.27  \\
 CaiT  &  0.66 \\
 \hline

\end{tabular}
  }
\end{minipage}
\hfill
\begin{minipage}[c]{0.58\textwidth}
\captionof{table}{\footnotesize {\textbf{CNNs with other explanation methods-}} We try 3 different explanation method for ResNet18 architecture and find that none of them can localize the patch correctly. Hence we do not see much difference in ASR.}
    \label{tab:interp_cnn}
\centering
\scalebox{1.0}{
\begin{tabular}{ |c||c|c|  }
 \hline
 &\multicolumn{1}{|c|}{Before Defense}&\multicolumn{1}{|c|}{After Defense} \\
 Method&   ASR  (\%) & ASR  (\%)  \\
 \hline
 GradCAM \cite{selvaraju2017grad} &41.80 &42.60 \\
 Score-CAM \cite{wang2020score} &41.80  & 42.18\\
FullGrad \cite{srinivas2019full} &41.80 & 43.20 \\
 \hline

\end{tabular}}

\end{minipage}

{\bf Using different interpretation algorithms for CNNs:} We also try different explanation algorithms for CNN architectures to ensure that our results are not biased towards a particular explanation method. Our defense results for 3 explanation methods \cite{selvaraju2017grad,srinivas2019full,wang2020score} on ResNet18 architecture is shown in Table \ref{tab:interp_cnn}. Note that the `Before Defense' results would be the same for all 3 rows. We find that non of the 3 explanation methods can help with localizing the patch. This show that CNNs cannot localize the patch due to the architecture, rather than the explanation algorithms.

\section{Conclusion}
\vspace{-.08in}
We show that existing threat models like BadNets and Hidden Trigger Backdoor Attacks are effective against Vision Transformers, thus showcasing the vulnerability of transformers to data poisoning. Interestingly, we find that in transformers, interpretation algorithms like GradRollout effectively highlights the trigger patch for a backdoor test-input. On the contrary, the trigger localization is not effective for CNNs. Based on this observation, we propose a test-time image blocking defense for backdoors. We empirically show that this defense effectively reduces the attack success rate for transformers. We hope that our results will encourage the community to study how Vision Transformers compare to CNNs in terms of adversarial robustness.

\textbf{Limitations:} In our threat model, the defender makes an assumption about the size of trigger patch encountered during test time. We also observe that the test-time image blocking causes a drop in the accuracy of clean test images. Additionally, by doing test time image blocking defence, our inference time increase by factor of $2$ since we need to forward twice per image.

\textbf{Societal Impact:} An adversary can use this form of attack to backdoor Vision Transformer models. Our results can be exploited by an adversary for unethical applications and backdoor transformer architectures. But we also propose a defense algorithm which greatly improves robustness to such adversaries. By highlighting such results, the research community can learn to build more robust architectures. On the other hand, it can also enable adversaries to construct better fooling algorithms.

{\bf Acknowledgment:}
This material is based upon work partially supported by the United States Air Force under Contract No. FA8750‐19‐C‐0098, funding from SAP SE, NSF grants 1845216 and 1920079, and also financial assistance award number 60NANB18D279 from U.S. Department of Commerce, National Institute of Standards and Technology. Any opinions, findings, and conclusions or recommendations expressed in this material are those of the authors and do not necessarily reflect the views of the United States Air Force, DARPA, or other funding agencies.


\bibliographystyle{unsrt}
\bibliography{main.bib}
\appendix

\section{Appendix}
\setcounter{table}{0}
\setcounter{figure}{0}
\renewcommand{\thetable}{A\arabic{table}}
\renewcommand{\thefigure}{A\arabic{figure}}

\maketitle

We report the results for each pair of categories in Tables A1-A4. Please refer to the caption for details. Also, Figure A1 shows some qualitative visualization.

\begin{figure}[!h]
  \begin{center}
  \scalebox{0.9}{
  \begin{tabular}{| c c c c c c|}
\hline  \footnotesize Original & \footnotesize{Patched} & \footnotesize{ResNet50} & \footnotesize{ResNet50 blocked} & \footnotesize{VIT-Base} &\footnotesize{VIT-Base blocked}\\
 \hline
\vspace{-.08in}
&&&&&\\

\begin{sideways} \scriptsize Target:Bee \end{sideways}
\includegraphics[width=.12\textwidth]{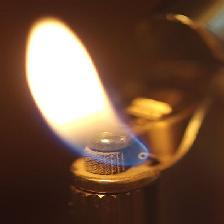}&
\includegraphics[width=.12\textwidth]{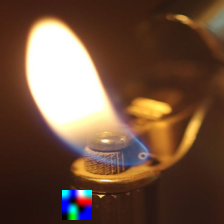}&
\includegraphics[width=.12\textwidth]{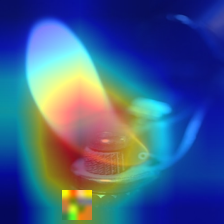}&
\includegraphics[width=.12\textwidth]{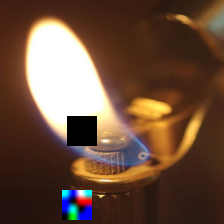}&
\includegraphics[width=.12\textwidth]{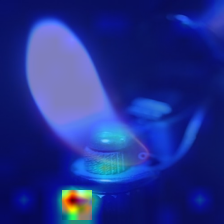}&
\includegraphics[width=.12\textwidth]{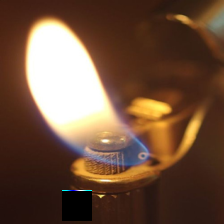}\\
\small Lighter & \small Bee  &  &\small Bee   & &\small Lighter  \\

\begin{sideways} \scriptsize Target:Partridge \end{sideways}
\includegraphics[width=.12\textwidth]{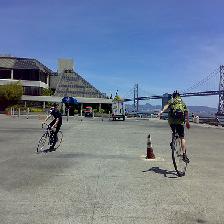}&
\includegraphics[width=.12\textwidth]{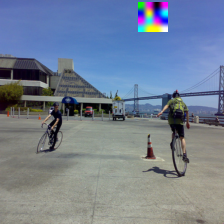}&
\includegraphics[width=.12\textwidth]{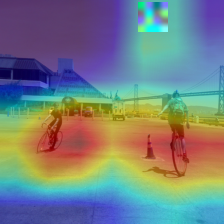}&
\includegraphics[width=.12\textwidth]{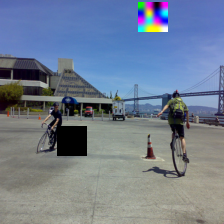}&
\includegraphics[width=.12\textwidth]{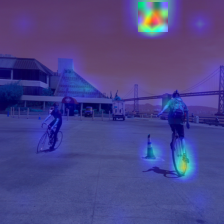}&
\includegraphics[width=.12\textwidth]{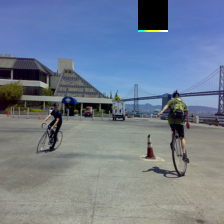}\\
\small Unicycle & \small Partridge  &  & \small Partridge  & &\small Unicycle  \\

\begin{sideways} \scriptsize Target:Deerhound \end{sideways}
\includegraphics[width=.12\textwidth]{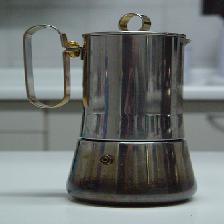}&
\includegraphics[width=.12\textwidth]{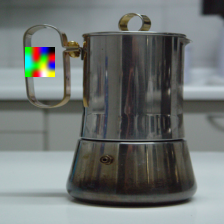}&
\includegraphics[width=.12\textwidth]{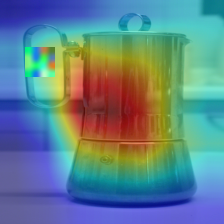}&
\includegraphics[width=.12\textwidth]{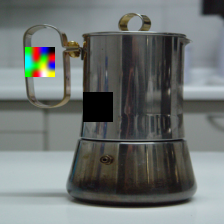}&
\includegraphics[width=.12\textwidth]{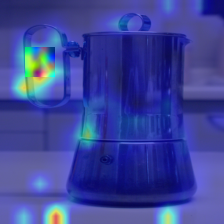}&
\includegraphics[width=.12\textwidth]{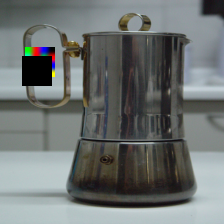}\\
\small Coffeepot & \small Deerhound  &  &  \small Deerhound  & &\small Coffeepot \\



\begin{sideways} \scriptsize Target:Toyshop \end{sideways}
\includegraphics[width=.12\textwidth]{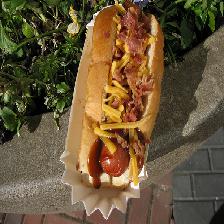}&
\includegraphics[width=.12\textwidth]{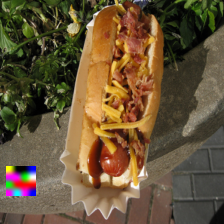}&
\includegraphics[width=.12\textwidth]{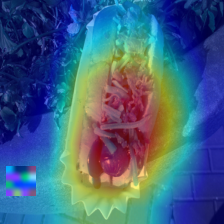}&
\includegraphics[width=.12\textwidth]{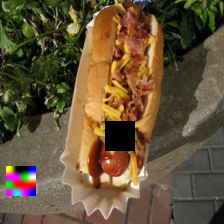}&
\includegraphics[width=.12\textwidth]{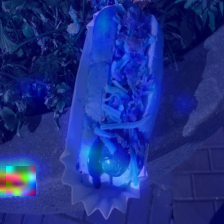}&
\includegraphics[width=.12\textwidth]{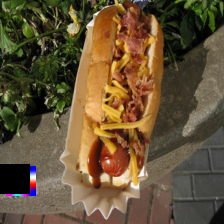}\\
\small Hotdog & \small Toyshop &  & \small Toyshop  & &\small Hotdog \\

\begin{sideways} \scriptsize Target:Tiger Beetle \end{sideways}
\includegraphics[width=.12\textwidth]{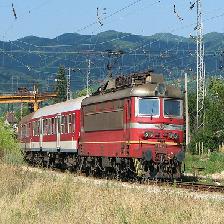}&
\includegraphics[width=.12\textwidth]{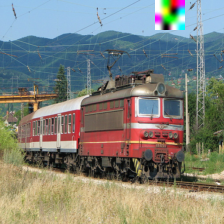}&
\includegraphics[width=.12\textwidth]{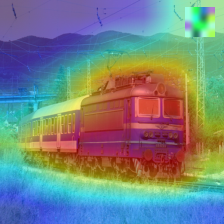}&
\includegraphics[width=.12\textwidth]{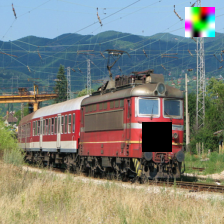}&
\includegraphics[width=.12\textwidth]{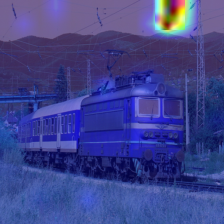}&
\includegraphics[width=.12\textwidth]{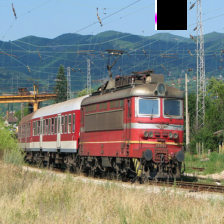}\\
\small Electric Locomotive & \small Tiger Beetle  &  &  \small Tiger Beetle & &\small Electric Locomotive \\


\begin{sideways} \centering \scriptsize Target:Plunger \end{sideways}
\includegraphics[width=.12\textwidth]{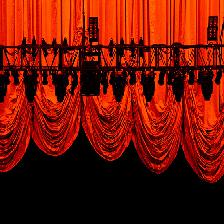}&
\includegraphics[width=.12\textwidth]{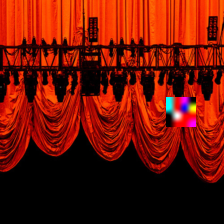}&
\includegraphics[width=.12\textwidth]{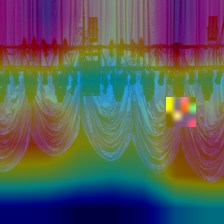}&
\includegraphics[width=.12\textwidth]{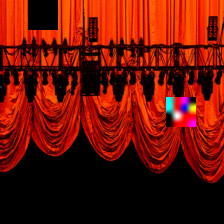}&
\includegraphics[width=.12\textwidth]{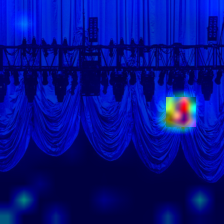}&
\includegraphics[width=.12\textwidth]{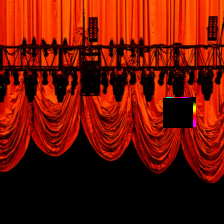}\\
\small Theater Curtain & \small Plunger &  & \small Plunger & &\small Theater Curtain \\

\begin{sideways} \centering \scriptsize Target:iPod \end{sideways}
\includegraphics[width=.12\textwidth]{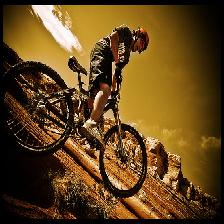}&
\includegraphics[width=.12\textwidth]{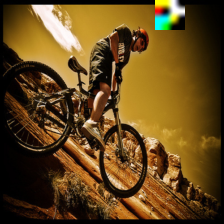}&
\includegraphics[width=.12\textwidth]{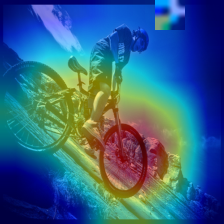}&
\includegraphics[width=.12\textwidth]{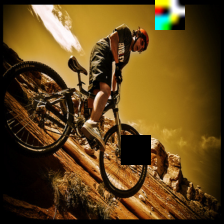}&
\includegraphics[width=.12\textwidth]{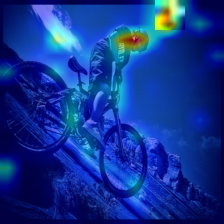}&
\includegraphics[width=.12\textwidth]{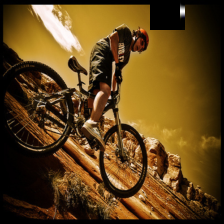}\\
\small Mountain Bike & \small iPod  &  & \small iPod & &\small Mountain Bike \\

\begin{sideways} \centering \scriptsize Target:Goblet \end{sideways}
\includegraphics[width=.12\textwidth]{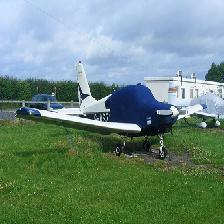}&
\includegraphics[width=.12\textwidth]{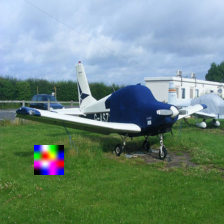}&
\includegraphics[width=.12\textwidth]{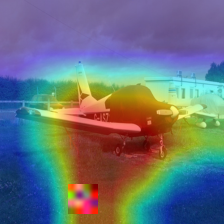}&
\includegraphics[width=.12\textwidth]{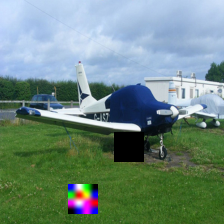}&
\includegraphics[width=.12\textwidth]{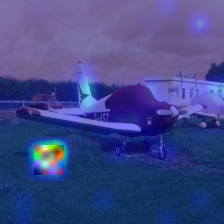}&
\includegraphics[width=.12\textwidth]{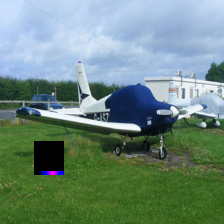}\\
\small Wing & \small Goblet  &  & \small Goblet & &\small Wing \\



\hline
\end{tabular}
}
\vspace{.02in}
  \caption{\footnotesize {\textbf{Image Blocking Defence-}} Qualitative results similar to Figure 3 of the main submission. Interestingly, we observe for the last case that the trigger does involve some heatmap for ResNet50, but since it is shared with the main object in the image and not the maximum region of the attention, the blocking mechanism does not work as expected. Attack was successful for all examples. }

\vspace{-.15in}
\label{supp_fig_patch_defense}
  \end{center}
\end{figure}

%
\begin{table}[!h]
  \caption{\footnotesize {\textbf{Results of Attack and Test time Defense-}}  To save space in the main article, we reported the results averaged over 10 random pairs of categories. In this table, we report the results for all pairs with ViT-Base architecture (similar to Table 1). The pairs of categories are the same random pairs used in HTBA [6]. Note that each pair of categories (each row) corresponds to a different attack task, so depending on the similarity of source and target categories, that attack may be easy or difficult. Hence, we do not expect a low standard deviation of ASR across these tasks. A similar large standard deviation was also reported in HTBA.}
    \label{tab:supp_test_time_defense}

  \scalebox{0.75}{
\begin{tabular}{ |c|c||c|c|c| c|c|c|  }
 \hline
 &&\multicolumn{3}{c|}{Attack}&\multicolumn{3}{c|}{Defense} \\
 \hline
 Source& Target& Val  & Source  &ASR &Val& Source  & ASR  \\
  &  &Accuracy (\%) &Accuracy (\%) &  (\%)  &Accuracy (\%) &Accuracy (\%) & (\%)  \\
 \hline

 Slot & Australian Terrier &79.02&92.00&56.00&76.94&92.00&6.00\\
 Lighter & Bee &79.06&66.00&58.00&76.95&70.00&28.00\\
 Theater Curtain & Plunger &78.96&90.00&82.00&76.95&78.00&20.00\\
 Unicycle & Partridge &79.04&92.00&70.00&76.99&70.00&14.00\\
 Mountain Bike & Ipod &79.04&78.00&68.00&76.86&66.00&30.00\\
 Coffeepot & Deerhound &79.04&64.00&52.00&76.93&66.00&16.00\\
 Can Opener & Cuckatoo &79.00&72.00&32.00&76.90&70.00&12.00\\
Hotdog & Toyshop &79.02&90.00&60.00&76.90&80.00&22.00\\
 Electric Locomotive & Tiger Beetle &79.04&88.00&84.00&76.99&92.00&6.00\\
 Wing & Goblet &79.18&42.00&52.00&76.98&48.00&10.00\\
 \hline
 \multicolumn{2}{|c||}{\textbf{Average}}&79.04 &77.4&61.4&76.94&73.2&16.4\\
 \multicolumn{2}{|c||}{\textbf{Standard Deviation}}&0.05 &16.5&15.40&0.04&13.10&8.47\\
 \hline

\end{tabular}
}
\label{tab:supp_vit_every_class}
\vspace{-0.15in}
\end{table}

\begin{table}[!h]
  \caption{\footnotesize {\textbf{Results of Attack and Test time Defense-}}  Similar to Table \ref{tab:supp_vit_every_class} for ResNet50 architecture.}
    \label{tab:supp_resnet50_every_class}

  \scalebox{0.75}{
\begin{tabular}{ |c|c||c|c|c| c|c|c|  }
 \hline
 &&\multicolumn{3}{c|}{Attack}&\multicolumn{3}{c|}{Defense} \\
 \hline
 Source& Target& Val  & Source  &ASR &Val& Source  & ASR  \\
  &  &Accuracy (\%) &Accuracy (\%) &  (\%)  &Accuracy (\%) &Accuracy (\%) & (\%)  \\
 \hline
 Slot & Australian Terrier &74.06&92.00&6.00&63.83&92.00&10.00\\
 Lighter & Bee &73.97&64.00&52.00&63.46&48.00&42.00\\
 Theater Curtain &Plunger &73.92&76.00&52.00&63.5&70.00&42.00\\
 Unicycle & Partridge &73.96&72.00&30.00&63.44&60.00&34.00\\
 Mountain Bike & Ipod &73.89&74.00&42.00&63.49&38.00&62.00\\
 Coffeepot & Deerhound &73.95&58.00&20.00&63.45&60.00&26.00\\
 Can Opener & Cuckatoo&73.88&70.00&18.00&63.59&60.00&22.00\\
Hotdog & Toyshop &73.84&78.00&60.00&63.41&36.00&60.00\\
 Electric Locomotive & Tiger Beetle &74.00&92.00&28.00&63.66&88.00&30.00\\
 Wing & Goblet &73.90&64.00&40.00&63.55&54.00&44.00\\
  \hline
 \multicolumn{2}{|c||}{\textbf{Average}}&73.94&74.00&34.8&63.538&60.6&37.2\\
 \multicolumn{2}{|c||}{\textbf{Standard Deviation}}&0.06&11.27&17.33&0.12&18.69&16.28\\
 \hline

\end{tabular}
}
\end{table}

\begin{table}[!h]
  \caption{\footnotesize {\textbf{Results of Attack and Test time Defense-}}  Similar to Table \ref{tab:supp_vit_every_class} for ResNet18 architecture.}
    \label{tab:supp_resnet18_every_class}

  \scalebox{0.75}{
\begin{tabular}{ |c|c||c|c|c| c|c|c|  }
 \hline
 &&\multicolumn{3}{c|}{Attack}&\multicolumn{3}{c|}{Defense} \\
 \hline
 Source& Target& Val  & Source  &ASR &Val& Source  & ASR  \\
  &  &Accuracy (\%) &Accuracy (\%) &  (\%)  &Accuracy (\%) &Accuracy (\%) & (\%)  \\
 \hline

 Slot & Australian Terrier &66.74&92.00&22.00&55.32&84.00&18.00\\
 Lighter & Bee &66.84&52.00&34.00&55.44&56.00&30.00\\
 Theater Curtain &Plunger &66.58&78.00&32.00&55.00&68.00&42.00\\
 Unicycle & Partridge &66.53&70.00&46.00&55.43&46.00&42.00\\
 Mountain Bike & Ipod &66.66&68.00&62.00&55.47&28.00&62.00\\
 Coffeepot & Deerhound &66.57&52.00&36.00&55.57&54.00&34.00\\
 Can Opener & Cuckatoo&66.75&58.00&42.00&55.64&48.00&42.00\\
Hotdog & Toyshop &66.67&70.00&42.00&55.16&48.00&64.00\\
 Electric Locomotive & Tiger Beetle &66.81&82.00&48.00&55.43&80.00&46.00\\
 Wing & Goblet &66.59&50.00&54.00&55.32&50.00&46.00\\
  \hline
 \multicolumn{2}{|c||}{\textbf{Average}}&66.67&67.2&41.80&55.37&56.2&42.60\\
 \multicolumn{2}{|c||}{\textbf{Standard Deviation}}&0.11&14.18&11.53&0.18&16.85&13.73\\

 \hline

\end{tabular}
}
\end{table}

\begin{table}[!h]
  \caption{\footnotesize {\textbf{Results of Attack and Test time Defense-}}  Similar to Table \ref{tab:supp_vit_every_class} for PatchConv architecture.}
    \label{tab:supp_patchconv_every_class}

  \scalebox{0.75}{
\begin{tabular}{ |c|c||c|c|c| c|c|c|  }
 \hline
 &&\multicolumn{3}{c|}{Attack}&\multicolumn{3}{c|}{Defense} \\
 \hline
 Source& Target& Val  & Source  &ASR &Val& Source  & ASR  \\
  &  &Accuracy (\%) &Accuracy (\%) &  (\%)  &Accuracy (\%) &Accuracy (\%) & (\%)  \\
 \hline

 Slot & Australian Terrier &80.19&94.00&58.00&75.96&96.00&2.00\\
 Lighter & Bee &80.67&84.00&64.00&76.31&70.00&24.00\\
 Theater Curtain &Plunger &80.23&84.00&42.00&75.97&78.00&18.00\\
 Unicycle & Partridge &80.25&88.00&32.00&75.97&76.00&16.00\\
 Mountain Bike & Ipod &80.28&86.00&28.00&75.93&74.00&18.00\\
 Coffeepot & Deerhound &80.19&68.00&34.00&76.11&66.00&8.00\\
 Can Opener & Cuckatoo&80.19&82.00&6.00&76.04&80.00&2.00\\
Hotdog & Toyshop &80.22&92.00&18.00&75.92&90.00&36.00\\
 Electric Locomotive & Tiger Beetle &80.16&88.00&80.00&75.95&88.00&4.00\\
 Wing & Goblet &80.18&42.00&22.00&75.93&46.00&16.00\\
 \hline
 \multicolumn{2}{|c||}{\textbf{Average}}&80.26&80.8&38.4&76.00&76.40&14.40\\
 \multicolumn{2}{|c||}{\textbf{Standard Deviation}}&0.15&15.35&22.82&0.12&14.13&10.78\\

 \hline

\end{tabular}
}
\end{table}

\end{document}